\documentclass{article}
\usepackage{ijcai24}

\usepackage{times}
\usepackage{soul}
\usepackage{url}
\usepackage[hidelinks]{hyperref}
\usepackage[utf8]{inputenc}
\usepackage[small]{caption}
\usepackage{graphicx}
\usepackage{amsmath}
\usepackage{amsthm}
\usepackage{booktabs}
\usepackage{algorithm}
\usepackage{algorithmic}
\usepackage[switch]{lineno}
\usepackage{multirow}
\usepackage{subcaption}
\usepackage{adjustbox}
\usepackage{wrapfig}
\usepackage{natbib}
\usepackage{colortbl}
\usepackage{multirow}
\usepackage{pifont}
\usepackage{enumitem}

\usepackage{xcolor}

\title{SignVTCL: Multi-Modal Continuous Sign Language Recognition Enhanced by Visual-Textual Contrastive Learning}

\author{
Hao Chen$^{1*}$\and
Jiaze Wang$^{1}$\thanks{Authors contributed equally}\and
Ziyu Guo$^1$\and
Jinpeng Li$^1$\and
Donghao Zhou$^1$\and
\\
Bian Wu$^2$\and
Chenyong Guan$^3$\and
Guangyong Chen$^2$ \thanks{Corresponding author}\and
Pheng-Ann Heng$^1$
\\
\affiliations
$^1$The Chinese University of Hong Kong\\
$^2$Zhejiang Lab\\
$^3$Gudsen Technology Co. Ltd\\
\emails
https://socialgoodai.github.io/,
}

\begin{document}

\maketitle

\begin{abstract}
Sign language recognition (SLR) plays a vital role in facilitating communication for the hearing-impaired community. SLR is a weakly supervised task where entire videos are annotated with glosses, making it challenging to identify the corresponding gloss within a video segment. Recent studies indicate that the main bottleneck in SLR is the insufficient training caused by the limited availability of large-scale datasets. To address this challenge, we present SignVTCL, a multi-modal continuous sign language recognition framework enhanced by visual-textual contrastive learning, which leverages the full potential of multi-modal data and the generalization ability of language model. SignVTCL integrates multi-modal data (video, keypoints, and optical flow) simultaneously to train a unified visual backbone, thereby yielding more robust visual representations. Furthermore, SignVTCL contains a visual-textual alignment approach incorporating gloss-level and sentence-level alignment to ensure precise correspondence between visual features and glosses at the level of individual glosses and sentence. Experimental results conducted on three datasets, Phoenix-2014, Phoenix-2014T, and CSL-Daily, demonstrate that SignVTCL achieves state-of-the-art results compared with previous methods.

\end{abstract}
\section{Introduction}

Videos provide a remarkably faithful representation of how humans consistently perceive the visual world. Consequently, the ability to comprehend videos is of paramount importance for artificial intelligence systems to acquire a profound understanding of the human world. This has positioned video understanding as the next frontier in the field of computer vision. Remarkable progress has been achieved in various video understanding tasks, such as classification \citep{ur2023use}, captioning \citep{sarto2023positive,yang2023vid2seq}, and action recognition \citep{xing2023svformer, ahn2023star}. However, when compared to these tasks, sign language recognition (SLR) presents ongoing challenges due to its inherent requirement for interpreting intricate and precise semantic information from videos. Consequently, addressing SLR using conventional video understanding methods is not a straightforward endeavor.
More importantly, the study of sign language is of great significance as it serves as a highly inclusive means of communication for the hearing-impaired community, effectively bridging the gap between deaf and hearing individuals. Therefore, investigating sign language recognition is a meaningful and impactful task.
In this work, we focus on studying continuous sign language recognition (CSLR) which aims to transcribe co-articulated sign videos into sign sequences on a gloss-by-gloss basis. In the subsequent discussion, we adopt the abbreviation SLR to refer to CSLR.


\begin{figure}[t]
    \centering
        \includegraphics[width=1.0\linewidth]{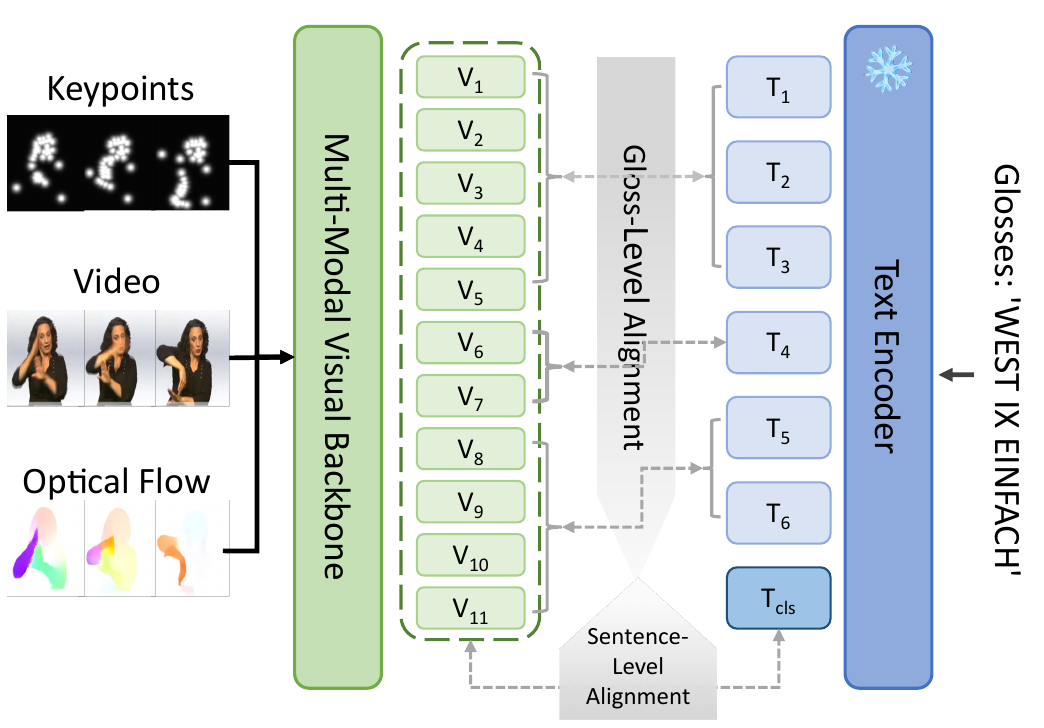}
    \caption{\textbf{The Overview of SignVTCL.} Three modalities of data are used to learn visual representations of sign language. The alignment at both the gloss and sentence level facilitates language-guided visual representation learning for boosting SLR capability.}
    \label{intro}
    \vspace{-8pt}
\end{figure}

Recently, many approaches \citep{cihan2017subunets,cheng2020fully,hu2023continuous} for SLR rely on computer vision techniques that extract features solely from video sequences. However, these methods often encounter challenges in effectively addressing the inherent complexities of sign language, such as the variations in signing styles among individuals and dynamic body parts movements. Additionally, the lack of large-scale, finely annotated sign datasets in the field of SLR often leads to insufficient training in these models.
Therefore, in this paper, we tackle these issues using multi-modal data with visual-textual contrastive learning.
Firstly, we focus on extracting different modal data, such as keypoints and optical flow from videos. These data sources offer diverse information to enhance the model's understanding of sign language, thereby improving sign language recognition capabilities. 
In parallel, we construct a multi-modal visual backbone that incorporates multi-modal fusion modules, allowing for simultaneous learning from multiple modalities. Additionally, we introduce a sign pyramid network (SPN) to effectively supervise the shallow layers of our backbone, facilitating the acquisition of meaningful visual representations.

Furthermore, inspired by the approach employed in CLIP~\citep{radford2021learning}, which leverages natural language supervision to learn image representations, we observe that SLR data inherently has a video-text pair structure. Hence, we posit that harnessing language-guided visual representations can serve as a valuable training task to enhance the performance of SLR.
Since SLR is a weakly supervised task that only annotates the sign video with textual glosses, our model faces the challenging task of determining the corresponding gloss for each video segment from diverse video segments.
Therefore, we propose a visual-textual alignment approach to align visual and textual feature embeddings from the gloss level and the sentence level. 
Gloss-Level alignment utilizes the dynamic time warping algorithm~\citep{berndt1994using} for calculating correspondence between frames and glosses then aligns the visual and textual features at the gloss level.
While sentence-level alignment aims to force the model to contain a holistic understanding of the semantic and contextual information within the sentence.
Thus the visual-textual alignment approach establishes a potential correspondence between visual signs and textual context.

Finally, by integrating the aforementioned techniques, we present \textbf{SignVTCL}, a multi-modal continuous sign language recognition framework enhanced by visual-textual contrastive learning, as shown in Figure \ref{intro}.
To validate the effectiveness of our proposed method, we conduct extensive experiments on Phoenix-2014 \citep{koller2015continuous}, Phoenix-2014T \citep{camgoz2018neural} and CSL-Daily \citep{zhou2021improving}. The results demonstrate that our approach achieves state-of-the-art performance in SLR.

Our main contributions can be summarized as follows:
\begin{itemize}[noitemsep,topsep=0pt,leftmargin=15pt]
\setlength{\itemsep}{0pt}
\item We are the first to effectively utilize video, keypoints, and optical flow modalities together to capture dynamic body parts movement information in SLR.
\item We present a visual-textual alignment approach to enhance the capability of SLR, which leverages textual data supervision from the gloss level and the sentence level.
\item The proposed SignVTCL achieves \textit{state-of-the-art} results across multiple benchmarks.
\end{itemize}

\section{Related Works}

\textbf{Sign Language Recognition.} Early efforts in SLR primarily relied on hand-crafted features \citep{han2009modelling,koller2015continuous} or Hidden Markov Model-based systems \citep{koller2016deep,koller2017re}. These approaches, while foundational, struggled to capture the inherent complexity and variability of sign language. In recent years, deep learning-based methodologies have ushered in a paradigm shift in the domain of SLR, which can be succinctly encapsulated within three pivotal phases: feature extraction, recognition, and alignment. Predominantly, 3D CNNs \citep{pu2019iterative,li2020tspnet,chen2022simple,chen2022two} have gained widespread adoption for feature extraction. Additionally, certain approaches \citep{hu2023continuous,min2021visual,zhou2021spatial,cui2019deep} opt to commence with a 2D CNN to extract frame-wise features before subsequently incorporating hybrid architectures composed of 1D CNNs and Long Short-Term Memory (LSTM) networks to capture temporal dependencies. Upon deriving features, classifiers can compute posterior probabilities to facilitate the recognition process. 
Finally, CTC loss \citep{graves2006connectionist} is widely used to find the proper alignment between clips and glosses to ensure an accurate training procedure. In addition to utilizing videos for SLR, keypoints \citep{chen2022two, chi2022infogcn} and optical flow \citep{cui2019deep} can be employed as auxiliary modal information to enhance the performance of SLR. Keypoints offer specific details about manual elements (handshape, palm orientation) and non-manual elements (facial expressions and movements of the body, head, mouth, eyes, and eyebrows). Optical flow, on the other hand, provides information about the movement of human body parts between consecutive frames of a video. In this study, we introduce a unified architecture that harnesses the information from three modalities (video, keypoints, optical flow) simultaneously to extract visual representations. By integrating these modalities, we aim to enhance the capability of SLR by capturing a more comprehensive understanding of sign language.

\noindent\textbf{Visual-Textual Contrastive Learning.}
Recently, there has been a growing trend in developing visual-textual approaches for visual problems. In these approaches, vision models are trained using free-form language supervision. For instance, CLIP \citep{radford2021learning} and ALIGN \citep{jia2021scaling} employ crossmodal contrastive learning on a large-scale dataset, using hundreds or even thousands of millions of image-text pairs. Inspired by this concept and recognizing the inherent video-text pair structure in SLR data, we have discovered that learning language-indicated visual representations from sign language videos is an effective approach to enhancing SLR performance. Few recent approaches have considered visual-textual contrastive learning in the field of sign language. \cite{zheng2023cvt} proposed a visual-textual transformation-based SLR framework that leverages the unique properties of autoencoders to implicitly align the visual and textual modalities. \cite{zhou2023gloss} integrated contrastive language-image pre-training with masked self-supervised learning to create pre-tasks that bridge the semantic gap between visual and textual representations. Both approaches utilize contrastive learning at the macro level, encompassing the entire videos and complete sentences within a mini-batch, necessitating an additional pre-training phase. However, we advance such methods by delving into the semantic alignment between distinct video segments and their corresponding glosses in the sentence, while eliminating the need for pre-training.

\begin{figure*}[t]
    \centering
        \includegraphics[width=\linewidth]{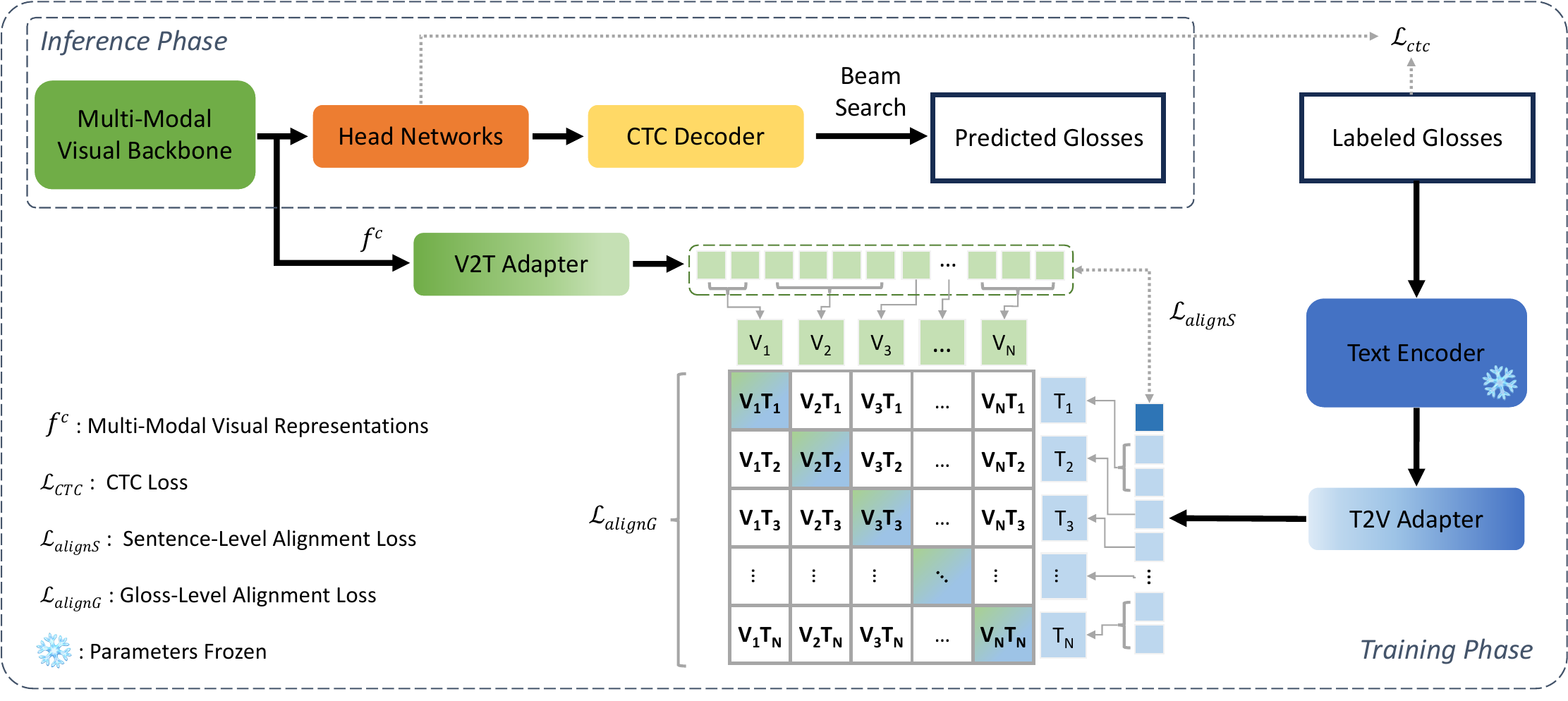}
    \caption{\textbf{The Pipeline of SignVTCL.} The multi-modal visual backbone aims to extract visual representations from three different modalities. These features are then passed through head networks for predicting frame-wise gloss probabilities. Simultaneously, we input labeled glosses into a frozen pretrained text encoder to obtain textual representations. Subsequently, the visual and textual representations are aligned within a joint multi-modal semantic space to supervise the multi-modal visual backbone using two adapters: the V2T adapter and the T2V adapter. During the inference phase, a CTC decoder is employed to generate glosses based on the predicted gloss probabilities.}
    \vspace{5pt}
    \label{fig:framework}
\end{figure*}

\section{Methods}
As shown in Figure \ref{fig:framework}, our SignVTCL consists of a multi-modal visual backbone, head networks, a text encoder, two feature adapters (V2T adapter and T2V adapter), and a CTC decoder. The multi-modal visual backbone is composed of a three-branch network, multi-modal fusion modules and sign pyramid networks. The task of SLR is to translate the input video $ x^v = \left\{x^v_{t} \right\}_{t=1}^{T} \in \mathcal{R}^{T\times H\times W\times 3}$ into a series of glosses $y = \left\{y_{i} \right\}_{i=1}^{N}$ to express a sentence, with $N$ denoting the length of the glosses sequence. 
In this section, we begin by introducing the multi-modal visual backbone. Then, we explain the utilization of a text encoder and two adapters to align visual and textual representations within a shared multi-modal semantic space. Finally, we delve into the training and inference processes of SignVTCL.

\subsection{Multi-Modal Visual Backbone}

\begin{figure}[t]
    \centering
        \includegraphics[width=0.88\linewidth]{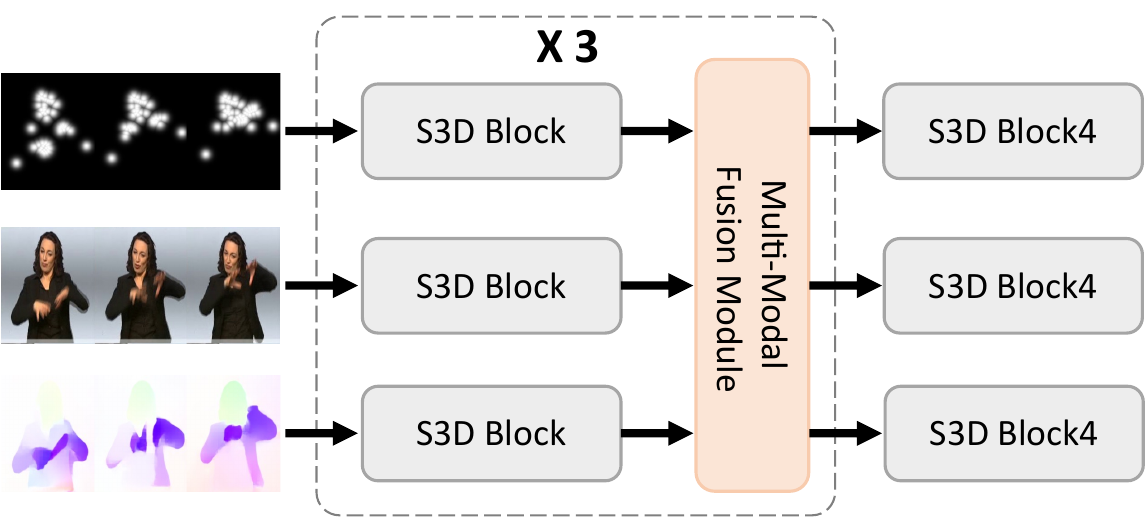}
    \vspace{4.5pt}
    \caption{\textbf{The Architecture of Three-Branch Network.} In each branch, the first four blocks of S3D serves as the backbone, providing the foundational architecture. Between each block, a multi-modal fusion module is incorporated to effectively merge information from different modalities.}
    \label{fig:backbone}
\end{figure}

\noindent \textbf{Three-Branch Network.}
To enhance the quality of our visual representation, we extracted keypoints and optical flow from the sign language video. These additional modalities are incorporated alongside the video itself to jointly train our visual backbone, forming a multi-modal training setup. 
We adopt the keypoints generated by HRNet \citep{wang2020deep} trained on COCO-WholeBody \citep{jin2020whole}, which contains 42 hand keypoints, 26 face keypoints, and 11 upper body keypoints, to model the keypoint sequences. To mitigate the sensitivity to noise, we opted to represent the keypoints using heatmaps instead of treating them as a set of coordinate points. Specifically, each coordinate point will be processed by a Gaussian function (explained in appendix) and become part of the heatmap. We treat each extracted keypoint as an independent channel input to the model. Therefore, the keypoints sequences can be represented as $ x^k = \left\{x^k_{t} \right\}_{t=1}^{T} \in \mathcal{R}^{T\times H\times W\times K}$, $K$ is the keypoints number.
Optical flow provides a dense motion representation, which enables the model to capture finer details of the signing motion and facilitates accurate recognition. By leveraging the capabilities of RAFT \citep{teed2020raft}, a deep network architecture explicitly crafted for extracting optical flow, we can proficiently acquire high-quality optical flow information from video data. We store the optical flow as a sequence of images. This format allows us to represent the optical flow information in a manner similar to video data, which can be represented as $ x^o = \left\{x^o_{t} \right\}_{t=1}^{T} \in \mathcal{R}^{T\times H\times W\times 3}$.
To facilitate effective learning of visual representations from the three modalities, we construct a network with three branches, as shown in Figure \ref{fig:backbone}. Each branch comprises the first four blocks of S3D \citep{xie2018rethinking}, a popular 3D convolutional network architecture commonly used for video understanding tasks. In the keypoint branch, we replace the first convolutional layer of the first block to accommodate the input format of keypoints.
Each branch processes input data into frame-wise features $f^{\{v,k,o\}} = \left\{f_{t}^{\{v,k,o\}} \right\}_{t=1}^{T/4} \in \mathcal{R}^{T/4 \times d_v}$, which will be individually fed into separate temporal heads in our head networks, where $d_v$ is the visual hidden dimension.

\noindent\textbf{Head Networks.}
We adopt separate temporal heads for the video branch, keypoint branch, and optical flow branch. 
To fully harness the potential of our three-branch architecture, we additionally combine the outputs of the video, keypoint, and optical flow branches. The combined representation  $f^c = \left\{f_{t}^c \right\}_{t=1}^{T/4} \in \mathcal{R}^{T/4 \times (3 \times d_v)}$ will be then fed into a joint temporal head, which shares the same architecture as the individual temporal heads of the video, keypoint, and optical flow. By merging the information from all three modalities in the joint temporal head, we aim to leverage the complementary strengths of each modality and capture a more comprehensive visual representation. Each temporal head is composed of a temporal linear layer, two temporal convolutional layers employing a kernel size of 3 and a stride of 1, and a linear layer as the classifier.
We forward $f^{\{v,k,o,c\}}$ into four separate temporal heads, which will output frame-wise gloss probabilities
$p^{\{v,k,o,c\}} = \left\{p_{t}^{\{v,k,o,c\}} \right\}_{t=1}^{T/4} \in \mathcal{R}^{T/4 \times C}$, $C$ is the size of the gloss vocabulary. Finally, $p^{\{v,k,o,c\}}$ will be used to compute CTC losses.

\noindent\textbf{Multi-Modal Fusion.}
To address the inefficiency of training the data from three modalities separately and only fusing them at the head networks, we introduce a multi-modal fusion module. This module is strategically positioned among the blocks of the S3D branch within our architecture, as depicted in Figure \ref{fig:backbone}. 
This module is implemented as a fully-connected MLP with two hidden layers, adhering to a straightforward design. The input of this module is the concatenation of hidden features from three modalities, while the output is the fused feature, which maintains the same size as the size of the hidden features from the three modalities. The fused feature will be added to the three input modal features to complete the fusion process.
We also studied other fusion methods, including attention-based, and convolution-based methods. Details of these fusion methods are placed in the appendix.

\begin{figure}[t]
    \centering
        \includegraphics[width=\linewidth]{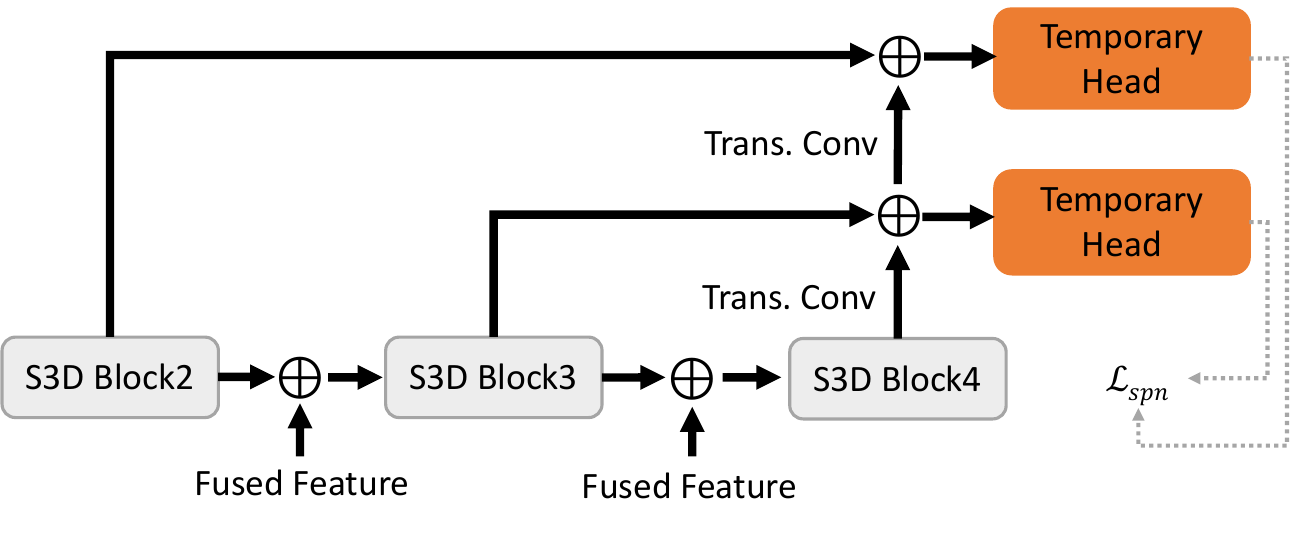}
    \caption{\textbf{The Architecture of Sign Pyramid Network (SPN).} To ensure the generation of meaningful representations, each branch is equipped with a SPN to supervise shallow layers. Transposed convolutions are employed to align the temporal and spatial dimensions of two feature maps, enabling element-wise addition between them.}
    \label{fig:spn}
    \vspace{4pt}
\end{figure}

\noindent\textbf{Sign Pyramid Network.}
To effectively supervise the shallow layers of our three-branch architecture for meaningful representations, we incorporate a sign pyramid network (SPN) with auxiliary supervision into our network. This approach builds upon previous research \citep{yang2020temporal} and involves a top-down pathway and lateral connections, as shown in Figure \ref{fig:spn}. To fuse features extracted by adjacent S3D blocks, we utilize an element-wise addition operation. Additionally, transposed convolutions are employed to match the temporal and spatial dimensions of the two feature maps before performing element-wise addition. Two separate temporal heads, following the same architecture as the temporal head in the head networks, are utilized to extract frame-wise gloss probabilities. Auxiliary supervision is provided by employing CTC losses. We employ three independent SPNs for the video, keypoint, and optical flow branches, ensuring comprehensive coverage across different modalities.

\subsection{Visual-Textual Alignment}
In this section, we introduce our visual-textual alignment approach to align visual and textual feature embeddings in a joint multi-modal semantic space. This approach establishes a potential correspondence between visual signs and textual context. 

\noindent\textbf{Text Encoder.}
To ensure the efficient encoding of textual data, the selection of a reliable text encoder holds paramount importance. Therefore, we opt for the encoder derived from mBART~\citep{liu2020multilingual}, which has undergone pre-training on CC25~\citep{liu2020multilingual}, an extensive multilingual corpus incorporating 25 languages. Given a sentence with $N$ glosses, it is first input to a tokenizer that converts the raw text to numbers, to generate $N_1$ tokens. Then, these tokens are fed into the text encoder to obtain high-dimensional semantic features $f^t = \left\{f_{n}^t \right\}_{n=1}^{N_1} \in \mathcal{R}^{N_1 \times d_t}$, where $d_t$ is the textual hidden dimension.


\noindent\textbf{Feature Adapters.}
To optimize the retention of both visual and pretrained textual knowledge, we freeze the pretrained text encoder and utilize it as a teacher model to supervise the learning of our multi-modal visual backbone. Inspired by ~\citep{gao2023clip,zhang2022tip}, we employ two lightweight adapters, a video-to-text (V2T) adapter and a text-to-video (T2V) adapter, to establish the connections between visual and textual features. 
The V2T adapter comprises two MLPs, with each MLP consisting of two hidden layers. One is dedicated to gloss-level alignment, while the other handles sentence-level alignment. The input of these two MLPs is the same visual feature $f^c$, and outputs are $f^{\{c_1, c_2\}} = \left\{f^{\{c_1, c_2\}}_t \right\}_{t=1}^{T/4} \in \mathcal{R}^{T/4 \times d_j}$, $d_j$ is the joint hidden dimension.
The T2V adapter is also implemented as an MLP with two hidden layers. We forward $f^t$ to this adapter to obtain $f^{t_1} = \left\{f_{n}^{t_1} \right\}_{n=1}^{N_1} \in \mathcal{R}^{N_1 \times d_j}$.
Thanks to these two adapters, we can align visual and textual representations within a unified multi-modal semantic space.

\begin{figure}[t]
    \centering
        \includegraphics[width=\linewidth]{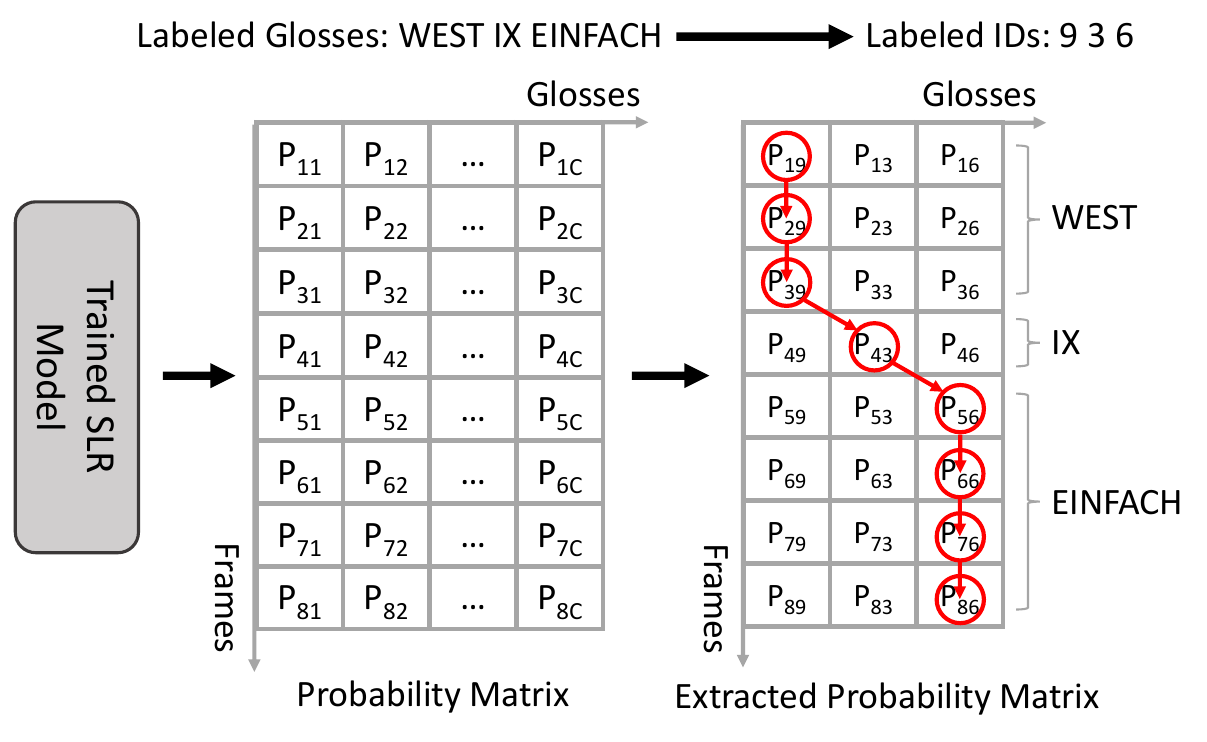}
    \caption{\textbf{An Example of Finding the Alignment of Gloss and Visual Features.} Assuming a vocabulary size of \textit{C} glosses, each gloss can be represented by an `ID' starting from 1. It is important to note that each row in the probability matrix should add up to 1. The guiding principle for determining the path with the highest probability is as follows: commencing from the top-left corner of the matrix, movement is restricted to either downward or towards the lower-right position. This constraint is imposed to maintain the correspondence between the order of video frames and labeled glosses.}
    \label{fig:align}
\end{figure}

\noindent\textbf{Gloss-Level Alignment.}
Gloss-Level alignment aims to integrate and align visual and textual features on the level of gloss. Due to the temporal characteristics of sign videos, glosses often have a one-to-many mapping to video frames while maintaining a consistent ordering. 
For textual features, token-wise features can be merged based on the labeled glosses, since we can ascertain the specific tokens to which each gloss is converted. We use the local average pooling to obtain $f^{t_{11}} = \left\{f_{n}^{t_{11}} \right\}_{n=1}^{N} \in \mathcal{R}^{N \times d_j}$ from $f^{t_1}$.
When it comes to visual features, a key challenge lies in assigning frame-wise features to the correct gloss. Drawing inspiration from the work of ~\citep{wei2023improving}, which utilizes the dynamic time warping (DTW) algorithm ~\citep{berndt1994using} for dictionary construction, we employ the DTW algorithm for aligning frame-wise features to glosses in hidden space. 
For details, first, a trained SLR model is utilized to generate a frame-wise probability matrix. From this matrix, the columns corresponding to the labeled glosses are extracted. Subsequently, a path is traced starting from the first position of the matrix, aiming to identify the path with the highest multiplication of probability values, as shown in Figure \ref{fig:align}. By following this path, the frame-wise visual features associated with each gloss can be determined.
Utilizing this knowledge, we can employ local average pooling to obtain $f^{c_{11}} = \left\{f_{n}^{c_{11}} \right\}_{n=1}^{N} \in \mathcal{R}^{N \times d_j}$ from $f^{c_1}$. 
Then the pair metrics can be calculated as:
\begin{equation}
\begin{aligned}
V2T_{pair} & = f^{c_{11}} \times (f^{t_{11}})^T ,\\
T2V_{pair} & = f^{t_{11}} \times (f^{c_{11}})^T ,
\end{aligned}
\end{equation}
where $\{V2T_{pair}, T2V_{pair}\} \in \mathcal{R}^{N \times N}$. Taking $V2T_{pair}$ as an example, $V2T_{pair}^{i,j}$ represents the similarity value of the $i$-th frame-wise visual feature and the $j$-th token-wise textual feature. The visual and textual features from the same glosses are positive samples, and the features from different glosses are negative samples.



\noindent\textbf{Sentence-Level Alignment.}
The objective of sentence-level alignment is to integrate and align visual features with textual features at the sentence level. In this process, a global textual feature $f^{t_{12}} \in \mathcal{R}^{1 \times d_j}$ that obtain from $f^{t_1}$ at the position of $<$EOS$>$ token is used to encapsulate the information of the entire sentence.
On the other hand, we employ a global pooling layer on $f^{c_2}$ to obtain the global visual feature $f^{c_{21}} \in \mathcal{R}^{1 \times d_j}$. Aligning these two features at the sentence level enables a holistic understanding of the semantic and contextual information conveyed by both visual and textual modalities within the sentence.


\subsection{Training and Inference}
\noindent\textbf{Multi-Modal SLR Losses.}
The multi-modal SLR losses consist of two parts. First, the CTC losses applied on the outputs of four separate temporal heads in our head networks $\mathcal{L}_{ctc}^v$, $\mathcal{L}_{ctc}^k$, $\mathcal{L}_{ctc}^{o}$ and $\mathcal{L}_{ctc}^j$. We sum them together to obtain $\mathcal{L}_{ctc}$. Second, the auxiliary CTC losses $\mathcal{L}_{spn}^v$, $\mathcal{L}_{spn}^k$, $\mathcal{L}_{spn}^o$ applied on the sign pyramid networks of three branches. We sum them together to obtain $\mathcal{L}_{spn}$. 


\noindent\textbf{Contrastive Alignment Losses.}
The contrastive alignment losses consist of two parts, the gloss-level alignment loss $\mathcal{L}_{alignG}$ and the sentence-level alignment loss $\mathcal{L}_{alignS}$. 
Given $V2T_{pair}$ and $T2V_{pair}$, the ground truth of them can be denoted as $G$, $G^{i,j}$ is equal to 1 when the $i$-th frame-wise visual feature and the $j$-th token-wise textual feature are corresponding, and otherwise 0. Therefore, the gloss-level alignment loss can be calculated as:
\begin{equation}
\begin{aligned}
L_{alginG}=\frac{1}{2}\Big(\text{CE\big(SoftMax}(V2T_{pair}\big) \times G_c, G\big) \\ +\ \text{CE\big(SoftMax}(T2V_{pair}) \times G_c, G\big)\Big)
\end{aligned}
\end{equation}
where $\text{CE}$ refers to the CrossEntroy loss, $G_c \in \mathcal{R}^{N \times 1}$ is a counter for counting the number of 1 in each line of $G$. 
The sentence-level alignment loss $\mathcal{L}_{alignS}$ is implemented by the KL divergence loss \citep{kullback1951information}.

\noindent\textbf{Total Loss.} We weighted sum the mentioned four losses to obtain the total training loss, which can be represented as: 
\begin{equation}
\begin{aligned}
\mathcal{L}_{t} = \lambda_{ctc} \mathcal{L}_{ctc} + \lambda_{spn} \mathcal{L}_{spn} + \lambda_{g} \mathcal{L}_{alignG} + \lambda_{s} \mathcal{L}_{alignS}.
\end{aligned}
\end{equation}

\noindent\textbf{Inference.}
In the inference stage, we employ a parameter-free CTC decoder to derive the final gloss predictions using the beam search algorithm with a beam width of 5. Specifically, the input is the average of frame-wise gloss probabilities $p^{\{v,k,o,c\}}$ that output from four separate temporal heads.

\begin{table*}[t]
    \centering
    \setlength{\tabcolsep}{3.5mm}{
    \begin{tabular}{l|cccccc}
        \toprule[1.5pt]
        \multirow{3}{*}{Method} & \multicolumn{4}{c}{Phoenix-2014} & \multicolumn{2}{c}{Phoenix-2014T}\\
        \cmidrule(lr){2-5}\cmidrule(lr){6-7}
        & \multicolumn{2}{c}{Dev (\%)} & \multicolumn{2}{c}{Test (\%)} & Dev (\%) & Test (\%)\\
        & DEL/INS & WER & DEL/INS & WER & WER & WER\\
        \midrule[0.5pt]
        SubUNets \citep{cihan2017subunets} & 14.6/4.0 &40.8 &  14.3/4.0 &40.7 &- &- \\
        CNN-LSTM-HMMs \citep{koller2019weakly} & - &26.0 & - &26.0  &24.1 &26.1\\
        DNF \citep{cui2019deep} &7.3/3.3 &23.1 & 6.7/3.3 &22.9 &- &-\\
        SFL \citep{niu2020stochastic} &7.9/6.5 &26.2  &7.5/6.3 &26.8 &- &- \\
        FCN \citep{cheng2020fully} &- &23.7 &- &23.9 &23.3 &25.1 \\
        Joint-SLRT \citep{camgoz2020sign} & - &- &- &- &24.6 &24.5 \\
        VAC \citep{min2021visual} &  7.9/2.5 &21.2& 8.4/2.6 &22.3 &- &- \\
        SignBT \citep{zhou2021improving} & - &- & - &- &22.7 &23.9 \\
        SMKD \citep{hao2021self} & 6.8/2.5 &20.8& 6.3/2.3 &21.0 &20.8 &22.4 \\
        STMC-R \citep{zhou2021spatial} &7.7/3.4 &21.1 &7.4/2.6 &20.7 &19.6 &21.0\\
        MMTLB \citep{chen2022simple} &- &- &- &- &21.9 &22.5 \\
        TLP \citep{hu2022temporal} & 6.3/2.8 & 19.7 & 6.1/2.9 &20.8 & 19.4 &21.2 \\
        C$^2$SLR \citep{zuo2022c2slr} & - &20.5 & - &20.4 &20.2 &20.4\\
        TwoStream-SLR \citep{chen2022two} & - &18.4 & - &18.8 &17.7 &19.3\\
        CorrNet \citep{hu2023continuous} &  \textbf{5.6}/2.8 &18.8 & \textbf{5.7}/2.3 &19.4 &18.9 &20.5 \\
        CVT-SLR \citep{zheng2023cvt} &  6.4/2.6 &19.8 & 6.1/2.3 &20.1 &19.4 &20.3 \\
        SEN \citep{hu2023self} & 5.8/2.6 & 19.5 & 7.3/4.0 &21.0 & 19.3 &20.7 \\
        \rowcolor{gray!20} \textbf{SignVTCL (Ours)} & 6.0/\textbf{2.4} & \textbf{17.3} & 5.9/\textbf{2.2} &\textbf{17.6} &\textbf{16.9} &\textbf{17.9}\\
        
        \bottomrule[1.5pt]
    \end{tabular}
    }
    \caption{\textbf{Comparison with Previous Methods on the Phoenix-2014 and Phoenix-2014T Datasets.} `DEL' and `INS' are average deletion and insertion rates.}
    \label{table: 1}
\end{table*}

\begin{table}[t]
    \vspace{5pt}
    \centering
    \setlength{\tabcolsep}{1mm}{
    \begin{tabular}{{l|cc}}
        \toprule[1.5pt]
        \multirow{2}*{Method} & \multicolumn{2}{c}{WER} \\
        \cmidrule(lr){2-3} & Dev (\%) & Test (\%)\\
        \midrule[0.5pt]
        SubUNets \citep{cihan2017subunets} & 41.4 &41.0\\
        LS-HAN \citep{huang2018video}  &39.0 &39.4\\
        FCN \citep{cheng2020fully}  &33.2 &33.5\\
        Joint-SLRT \citep{camgoz2020sign}  &33.1 &32.0\\
        SignBT \citep{zhou2021improving} &33.2 &33.2\\
        TwoStream-SLR \citep{chen2022two}  & 25.4 &25.3\\
        CorrNet \citep{hu2023continuous}  & 30.6 &30.1 \\
        SEN \citep{hu2023self}  & 31.1 &30.7 \\
        \rowcolor{gray!20} \textbf{SignVTCL (Ours)}  & \textbf{24.3} &\textbf{24.1} \\
        \bottomrule[1.5pt]
    \end{tabular}
    }
    \caption{\textbf{Comparison with Previous Methods on the CSL-Daily Dataset.}}
    \label{table: 2}
    \vspace{-10pt}
\end{table}

\section{Experiments}

\subsection{Experimental Setup}

\textbf{Datasets.}
\textbf{Phoenix-2014} \citep{koller2015continuous} and \textbf{Phoenix-2014T} \citep{camgoz2018neural} are two German sign language datasets widely used in the field of SLR. The Phoenix-2014 dataset consists of 5672 training, 540 development, and 629 testing samples, with a vocabulary size of 1295 for glosses. On the other hand, Phoenix-2014T is an extension of Phoenix-2014, containing 7096 training, 519 development, and 642 testing samples and has a vocabulary size of 1085 for glosses.
\textbf{CSL-Daily} \citep{zhou2021improving} is a recently released large-scale Chinese sign language dataset. It comprises 18401 training, 1077 development, and 1176 testing video samples. The dataset captures the performances of ten different signers and covers various topics such as family life, medical care, and school life. CSL-Daily features a vocabulary size of 2000 for glosses.


\noindent\textbf{Evaluation Metrics.}
The Word Error Rate (WER) stands as the predominant metric for assessing the performance of SLR. It quantifies the essential insertions (\#ins), substitutions (\#sub),  and deletions (\#del) required to align predicted sentences with their corresponding reference sentences (\#reference). The lower WER, the better accuracy.
\begin{equation} 
     \text{WER} = \frac{\#ins + \#sub + \#del}{\#reference}.
\end{equation}

\noindent\textbf{Implementation Details.} In both the Phoenix-2014 and Phoenix-2014T datasets, data from three modalities are resized and cropped to dimensions of 224 × 224, while in the CSL-Daily dataset, a crop size of 320 × 320 is applied. During the training phase, data augmentations are applied, consisting of spatial cropping within the range of [0.7-1.0] and frame-rate augmentation spanning [×0.5-×1.5]. For our network training strategy, we implement a cosine annealing schedule spanning 60 epochs. We utilize the Adam optimizer with a weight decay of  $1e^{-3}$ and set the initial learning rate to $1e^{-3}$. We train our models on 4 Nvidia A100 GPUs.

\subsection{Comparison with State-of-the-art Methods}
\textbf{Evaluation on Phoenix-2014 and Phoenix-2014T.}
As illustrated in Table \ref{table: 1}, we conducted a comparative analysis between our proposed SignVTCL and the existing state-of-the-art methods. Our approach outperformed all the other methods, establishing new state-of-the-art results on both the Phoenix-2014 and Phoenix-2014T datasets. 
The WER scores achieved by SignVTCL on the development set of Phoenix-2014 and Phoenix-2014T are 17.3\% and 16.9\%, respectively. While on the test set, the optimal WER scores achieved by SignVTCL outperform the previous best method by 1.2\% on and 1.4\%, respectively. 
Also, we can see SignVTCL achieves a very low percentage of deletion and insertion compared with other methods.

\noindent\textbf{Evaluation on CSL-Daily.} In Table \ref{table: 2}, we present a comparative analysis between our SignVTCL and the previous state-of-the-art methods on the CSL-Daily dataset. Our model exhibits a notable reduction in WER by 1.1\% on the development set and 1.2\% on the testing set when compared to TwoStream-SLR, showcasing its superior performance. 

\subsection{Ablation Studies}
\noindent\textbf{Effectiveness of Each Encoder and Fusion Module in SignVTCL.}
In Table \ref{table: 3}, we assessed the performance of SLR by individually employing the video encoder, keypoint encoder, and optical flow encoder. Notably, the video encoder yielded the most favorable outcomes, while the flow encoder demonstrated comparatively less effectiveness. Specifically, on the development and testing sets, WER of 20.5\% and 20.9\%, respectively, were achieved using the video encoder, while the flow encoder resulted in WER of 39.4\% and 38.4\%, respectively. Upon combining these three encoders, a substantial enhancement in sign language recognition performance was observed, showcasing a significant reduction in WER compared to using any single encoder in isolation. 
While further employing our fusion modules, a notable improvement was observed, resulting in WER reductions of 1.1\% on the development set and 1.0\% on the test set.
It is worth noting that in this set of ablation experiments, we intentionally excluded the visual-textual alignment approach.

\begin{figure}[t]
    \centering
        \includegraphics[width=\linewidth]{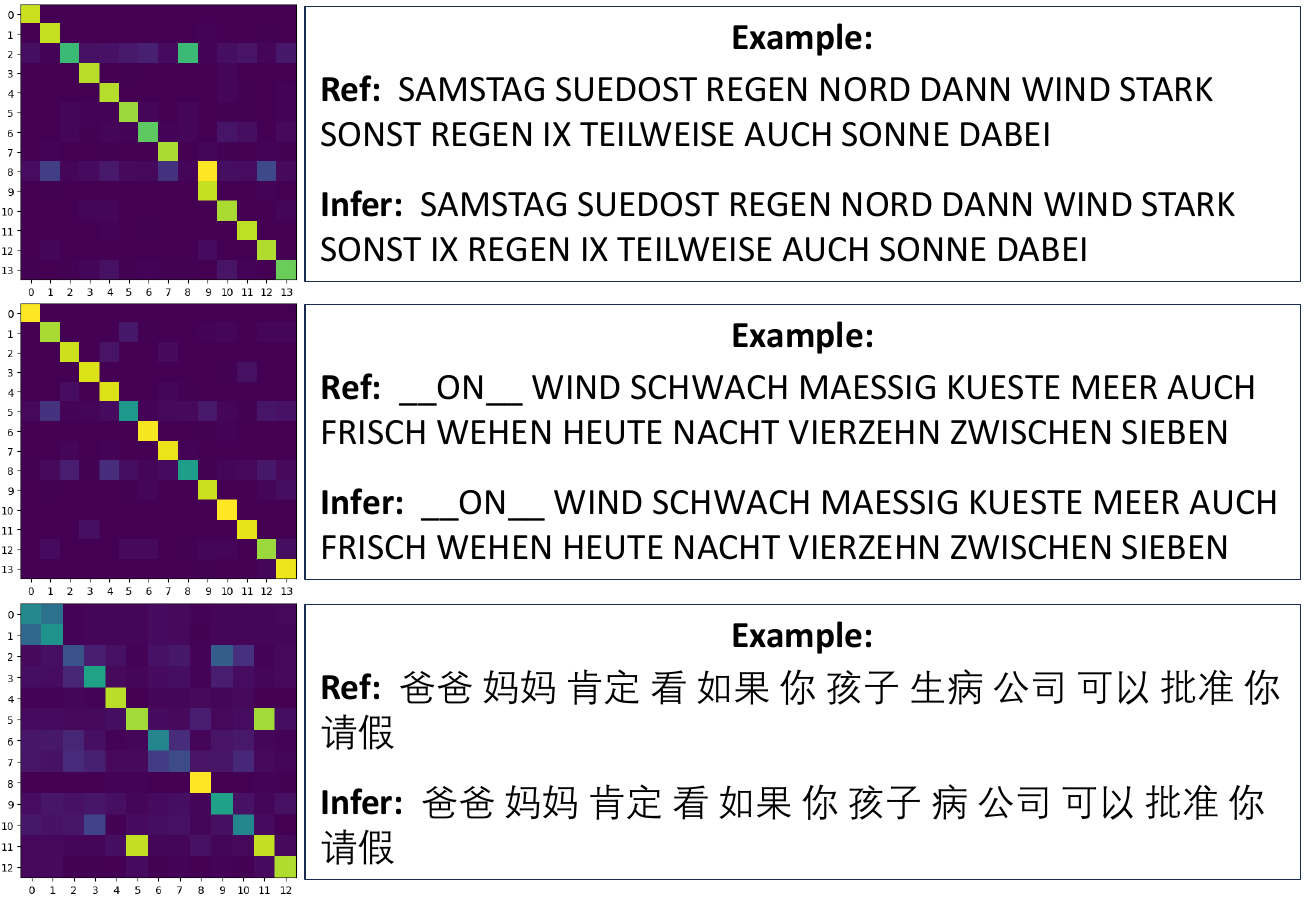}
    \caption{\textbf{Visualization of Gloss-Level Alignment Similarity Matrix and Predicted Glosses.} Each value within the similarity matrix corresponds to the degree of similarity observed between visual and textual representations for different glosses.}
    \label{fig:vis}
\end{figure}

\begin{table}[!]
    \centering
    \setlength{\tabcolsep}{3.5mm}{
    \begin{tabular}{{cccc|cc}}
        \toprule[1.5pt]
        \multirow{2}*{VE}& \multirow{2}*{KE} & \multirow{2}*{OE} & \multirow{2}*{FM} & \multicolumn{2}{c}{WER} \\
        \cmidrule(lr){5-6} & & & & Dev (\%) &Test (\%)\\
        \midrule[0.5pt]
         \ding{51}&  &  & & 20.5& 20.9\\
         &  \ding{51} &  & & 27.3& 26.5\\
         &   &  \ding{51} & & 39.4& 38.4\\
        \ding{51} & \ding{51} &  & & 19.9& 20.1\\
        \ding{51} &  \ding{51} & \ding{51} & & 19.4& 19.6\\
        \rowcolor{gray!20} \ding{51} &  \ding{51} & \ding{51} & \ding{51} & \textbf{18.3}& \textbf{18.6}\\
        \bottomrule[1.5pt]
    \end{tabular}
    }
    \caption{\textbf{Ablation Study for the Effectiveness of Each Encoder and Fusion Module on the Phoenix-2014 Dataset.} `VE', `KE', `OE' are video encoder, keypoint encoder, and optical flow encoder, respectively. `FM' denotes the fusion module.} 
    \label{table: 3}
\end{table}

\noindent\textbf{Effectiveness of Each Training Loss.} In Table \ref{table: 5}, when we solely employ CTC losses on the outputs of four separate temporal heads in our head networks, the optimal WER scores achieved by SignVTCL are 18.3\% and 18.6\%. In order to encompass glosses spanning various temporal extents and encourage intermediate layers to acquire more semantically meaningful features, we design SPNs to craft auxiliary CTC losses. The results in the second row of Table \ref{table: 5} show its effectiveness.
Additionally, we study the gloss-level alignment loss and sentence-level alignment loss respectively, and the results are shown in the third row and fourth row of this table.
When we combined these losses together for training SignVTCL, a final optimization result was obtained, 17.3\% and 17.6\% WERs on the development and test sets respectively.


\begin{table}[t]
    \centering
    \setlength{\tabcolsep}{3mm}{
    \begin{tabular}{{cc|cc}}
        \toprule[1.5pt]
        \multirow{2}*{\#} & \multirow{2}*{Training Losses} & \multicolumn{2}{c}{WER} \\
        \cmidrule(lr){3-4}& & Dev (\%) &Test (\%)\\
        \midrule[0.5pt]
         1 &  $\mathcal{L}_{ctc}$ &  18.3& 18.6\\
         2 &  1 + $\mathcal{L}_{spn}$ &  18.0& 18.3\\
         3 &  2 + $\mathcal{L}_{alignG}$ & 17.5 & 17.8\\
         4 &  2 + $\mathcal{L}_{alignS}$ &  17.7& 17.9\\
         \rowcolor{gray!20} 5 &  3 + $\mathcal{L}_{alignS}$ &  \textbf{17.3}& \textbf{17.6}\\
        \bottomrule[1.5pt]
    \end{tabular}
    }
    \caption{\textbf{Ablation Study for the Effectiveness of Each Training Loss on the Phoenix-2014 Dataset.}}
    \label{table: 5}
\end{table}

\subsection{Visualizations}
To visualize the similarity matrix and predicted glosses, we randomly selected an example from each of the three datasets, Phoenix-2014T, Phoenix-2014, and CSL-Daily, in Figure \ref{fig:vis}. Based on the three predicted glosses compared to their reference glosses, it is apparent that the reference and inferred glosses exhibit a high degree of similarity, leading to relatively perfect performance. Notably, the highlighted regions in the alignment matrices predominantly concentrate in proximity to the diagonal. This concentration serves as a clear indication that our gloss-level alignment plays a pivotal role in efficiently aligning visual and textual features, thereby enhancing SLR performance. More visualization examples can be seen in the appendix.

\section{Conclusion}
In this paper, we propose SignVTCL, a multi-modal continuous sign language recognition framework that integrates video, keypoints, and optical flow modalities together to learn visual representations. By combining these modalities, SignVTCL captures the intricate hand movements and dynamic body parts movements involved in sign language, enhancing the model's understanding of sign language and improving recognition capabilities.
Furthermore, we introduce a visual-textual alignment approach that aligns visual and textual feature embeddings at both the gloss and sentence levels, which establishes a meaningful and precise correspondence between visual signs and textual context, enhancing the performance of SLR. Extensive experiments on multiple datasets demonstrate the effectiveness of our SignVTCL in achieving state-of-the-art performance. We expect the proposed SignVTCL can inspire other works to explore multi-modality learning and contrastive learning in video understanding tasks.



\bibliographystyle{named}
\bibliography{ijcai24}

\begin{thebibliography}{}

\bibitem[\protect\citeauthoryear{Ahn \bgroup \em et al.\egroup
  }{2023}]{ahn2023star}
Dasom Ahn, Sangwon Kim, Hyunsu Hong, and Byoung~Chul Ko.
\newblock Star-transformer: a spatio-temporal cross attention transformer for
  human action recognition.
\newblock In {\em Proceedings of the IEEE/CVF Winter Conference on Applications
  of Computer Vision}, pages 3330--3339, 2023.

\bibitem[\protect\citeauthoryear{Berndt and Clifford}{1994}]{berndt1994using}
Donald~J Berndt and James Clifford.
\newblock Using dynamic time warping to find patterns in time series.
\newblock In {\em Proceedings of the 3rd international conference on knowledge
  discovery and data mining}, pages 359--370, 1994.

\bibitem[\protect\citeauthoryear{Camgoz \bgroup \em et al.\egroup
  }{2018}]{camgoz2018neural}
Necati~Cihan Camgoz, Simon Hadfield, Oscar Koller, Hermann Ney, and Richard
  Bowden.
\newblock Neural sign language translation.
\newblock In {\em Proceedings of the IEEE conference on computer vision and
  pattern recognition}, pages 7784--7793, 2018.

\bibitem[\protect\citeauthoryear{Camgoz \bgroup \em et al.\egroup
  }{2020}]{camgoz2020sign}
Necati~Cihan Camgoz, Oscar Koller, Simon Hadfield, and Richard Bowden.
\newblock Sign language transformers: Joint end-to-end sign language
  recognition and translation.
\newblock In {\em Proceedings of the IEEE/CVF conference on computer vision and
  pattern recognition}, pages 10023--10033, 2020.

\bibitem[\protect\citeauthoryear{Chen \bgroup \em et al.\egroup
  }{2022a}]{chen2022simple}
Yutong Chen, Fangyun Wei, Xiao Sun, Zhirong Wu, and Stephen Lin.
\newblock A simple multi-modality transfer learning baseline for sign language
  translation.
\newblock In {\em Proceedings of the IEEE/CVF Conference on Computer Vision and
  Pattern Recognition}, pages 5120--5130, 2022.

\bibitem[\protect\citeauthoryear{Chen \bgroup \em et al.\egroup
  }{2022b}]{chen2022two}
Yutong Chen, Ronglai Zuo, Fangyun Wei, Yu~Wu, Shujie Liu, and Brian Mak.
\newblock Two-stream network for sign language recognition and translation.
\newblock {\em Advances in Neural Information Processing Systems},
  35:17043--17056, 2022.

\bibitem[\protect\citeauthoryear{Cheng \bgroup \em et al.\egroup
  }{2020}]{cheng2020fully}
Ka~Leong Cheng, Zhaoyang Yang, Qifeng Chen, and Yu-Wing Tai.
\newblock Fully convolutional networks for continuous sign language
  recognition.
\newblock In {\em Computer Vision--ECCV 2020: 16th European Conference,
  Glasgow, UK, August 23--28, 2020, Proceedings, Part XXIV 16}, pages 697--714.
  Springer, 2020.

\bibitem[\protect\citeauthoryear{Chi \bgroup \em et al.\egroup
  }{2022}]{chi2022infogcn}
Hyung-gun Chi, Myoung~Hoon Ha, Seunggeun Chi, Sang~Wan Lee, Qixing Huang, and
  Karthik Ramani.
\newblock Infogcn: Representation learning for human skeleton-based action
  recognition.
\newblock In {\em Proceedings of the IEEE/CVF Conference on Computer Vision and
  Pattern Recognition}, pages 20186--20196, 2022.

\bibitem[\protect\citeauthoryear{Cihan~Camgoz \bgroup \em et al.\egroup
  }{2017}]{cihan2017subunets}
Necati Cihan~Camgoz, Simon Hadfield, Oscar Koller, and Richard Bowden.
\newblock Subunets: End-to-end hand shape and continuous sign language
  recognition.
\newblock In {\em Proceedings of the IEEE international conference on computer
  vision}, pages 3056--3065, 2017.

\bibitem[\protect\citeauthoryear{Cui \bgroup \em et al.\egroup
  }{2019}]{cui2019deep}
Runpeng Cui, Hu~Liu, and Changshui Zhang.
\newblock A deep neural framework for continuous sign language recognition by
  iterative training.
\newblock {\em IEEE Transactions on Multimedia}, 21(7):1880--1891, 2019.

\bibitem[\protect\citeauthoryear{Gao \bgroup \em et al.\egroup
  }{2023}]{gao2023clip}
Peng Gao, Shijie Geng, Renrui Zhang, Teli Ma, Rongyao Fang, Yongfeng Zhang,
  Hongsheng Li, and Yu~Qiao.
\newblock Clip-adapter: Better vision-language models with feature adapters.
\newblock {\em International Journal of Computer Vision}, pages 1--15, 2023.

\bibitem[\protect\citeauthoryear{Graves \bgroup \em et al.\egroup
  }{2006}]{graves2006connectionist}
Alex Graves, Santiago Fern{\'a}ndez, Faustino Gomez, and J{\"u}rgen
  Schmidhuber.
\newblock Connectionist temporal classification: labelling unsegmented sequence
  data with recurrent neural networks.
\newblock In {\em Proceedings of the 23rd international conference on Machine
  learning}, pages 369--376, 2006.

\bibitem[\protect\citeauthoryear{Han \bgroup \em et al.\egroup
  }{2009}]{han2009modelling}
Junwei Han, George Awad, and Alistair Sutherland.
\newblock Modelling and segmenting subunits for sign language recognition based
  on hand motion analysis.
\newblock {\em Pattern Recognition Letters}, 30(6):623--633, 2009.

\bibitem[\protect\citeauthoryear{Hao \bgroup \em et al.\egroup
  }{2021}]{hao2021self}
Aiming Hao, Yuecong Min, and Xilin Chen.
\newblock Self-mutual distillation learning for continuous sign language
  recognition.
\newblock In {\em Proceedings of the IEEE/CVF International Conference on
  Computer Vision}, pages 11303--11312, 2021.

\bibitem[\protect\citeauthoryear{Hu \bgroup \em et al.\egroup
  }{2022}]{hu2022temporal}
Lianyu Hu, Liqing Gao, Zekang Liu, and Wei Feng.
\newblock Temporal lift pooling for continuous sign language recognition.
\newblock In {\em European Conference on Computer Vision}, pages 511--527.
  Springer, 2022.

\bibitem[\protect\citeauthoryear{Hu \bgroup \em et al.\egroup
  }{2023a}]{hu2023continuous}
Lianyu Hu, Liqing Gao, Zekang Liu, and Wei Feng.
\newblock Continuous sign language recognition with correlation network.
\newblock In {\em Proceedings of the IEEE/CVF Conference on Computer Vision and
  Pattern Recognition}, pages 2529--2539, 2023.

\bibitem[\protect\citeauthoryear{Hu \bgroup \em et al.\egroup
  }{2023b}]{hu2023self}
Lianyu Hu, Liqing Gao, Zekang Liu, and Wei Feng.
\newblock Self-emphasizing network for continuous sign language recognition.
\newblock In {\em Proceedings of the AAAI Conference on Artificial
  Intelligence}, volume~37, pages 854--862, 2023.

\bibitem[\protect\citeauthoryear{Huang \bgroup \em et al.\egroup
  }{2018}]{huang2018video}
Jie Huang, Wengang Zhou, Qilin Zhang, Houqiang Li, and Weiping Li.
\newblock Video-based sign language recognition without temporal segmentation.
\newblock In {\em Proceedings of the AAAI Conference on Artificial
  Intelligence}, volume~32, 2018.

\bibitem[\protect\citeauthoryear{Jia \bgroup \em et al.\egroup
  }{2021}]{jia2021scaling}
Chao Jia, Yinfei Yang, Ye~Xia, Yi-Ting Chen, Zarana Parekh, Hieu Pham, Quoc Le,
  Yun-Hsuan Sung, Zhen Li, and Tom Duerig.
\newblock Scaling up visual and vision-language representation learning with
  noisy text supervision.
\newblock In {\em International conference on machine learning}, pages
  4904--4916. PMLR, 2021.

\bibitem[\protect\citeauthoryear{Jin \bgroup \em et al.\egroup
  }{2020}]{jin2020whole}
Sheng Jin, Lumin Xu, Jin Xu, Can Wang, Wentao Liu, Chen Qian, Wanli Ouyang, and
  Ping Luo.
\newblock Whole-body human pose estimation in the wild.
\newblock In {\em Computer Vision--ECCV 2020: 16th European Conference,
  Glasgow, UK, August 23--28, 2020, Proceedings, Part IX 16}, pages 196--214.
  Springer, 2020.

\bibitem[\protect\citeauthoryear{Koller \bgroup \em et al.\egroup
  }{2015}]{koller2015continuous}
Oscar Koller, Jens Forster, and Hermann Ney.
\newblock Continuous sign language recognition: Towards large vocabulary
  statistical recognition systems handling multiple signers.
\newblock {\em Computer Vision and Image Understanding}, 141:108--125, 2015.

\bibitem[\protect\citeauthoryear{Koller \bgroup \em et al.\egroup
  }{2016}]{koller2016deep}
Oscar Koller, O~Zargaran, Hermann Ney, and Richard Bowden.
\newblock Deep sign: Hybrid cnn-hmm for continuous sign language recognition.
\newblock In {\em Proceedings of the British Machine Vision Conference 2016},
  2016.

\bibitem[\protect\citeauthoryear{Koller \bgroup \em et al.\egroup
  }{2017}]{koller2017re}
Oscar Koller, Sepehr Zargaran, and Hermann Ney.
\newblock Re-sign: Re-aligned end-to-end sequence modelling with deep recurrent
  cnn-hmms.
\newblock In {\em Proceedings of the IEEE conference on computer vision and
  pattern recognition}, pages 4297--4305, 2017.

\bibitem[\protect\citeauthoryear{Koller \bgroup \em et al.\egroup
  }{2019}]{koller2019weakly}
Oscar Koller, Necati~Cihan Camgoz, Hermann Ney, and Richard Bowden.
\newblock Weakly supervised learning with multi-stream cnn-lstm-hmms to
  discover sequential parallelism in sign language videos.
\newblock {\em IEEE transactions on pattern analysis and machine intelligence},
  42(9):2306--2320, 2019.

\bibitem[\protect\citeauthoryear{Kullback and
  Leibler}{1951}]{kullback1951information}
Solomon Kullback and Richard~A Leibler.
\newblock On information and sufficiency.
\newblock {\em The annals of mathematical statistics}, 22(1):79--86, 1951.

\bibitem[\protect\citeauthoryear{Li \bgroup \em et al.\egroup
  }{2020}]{li2020tspnet}
Dongxu Li, Chenchen Xu, Xin Yu, Kaihao Zhang, Benjamin Swift, Hanna Suominen,
  and Hongdong Li.
\newblock Tspnet: Hierarchical feature learning via temporal semantic pyramid
  for sign language translation.
\newblock {\em Advances in Neural Information Processing Systems},
  33:12034--12045, 2020.

\bibitem[\protect\citeauthoryear{Liu \bgroup \em et al.\egroup
  }{2020}]{liu2020multilingual}
Yinhan Liu, Jiatao Gu, Naman Goyal, Xian Li, Sergey Edunov, Marjan
  Ghazvininejad, Mike Lewis, and Luke Zettlemoyer.
\newblock Multilingual denoising pre-training for neural machine translation.
\newblock {\em Transactions of the Association for Computational Linguistics},
  8:726--742, 2020.

\bibitem[\protect\citeauthoryear{Min \bgroup \em et al.\egroup
  }{2021}]{min2021visual}
Yuecong Min, Aiming Hao, Xiujuan Chai, and Xilin Chen.
\newblock Visual alignment constraint for continuous sign language recognition.
\newblock In {\em Proceedings of the IEEE/CVF International Conference on
  Computer Vision}, pages 11542--11551, 2021.

\bibitem[\protect\citeauthoryear{Niu and Mak}{2020}]{niu2020stochastic}
Zhe Niu and Brian Mak.
\newblock Stochastic fine-grained labeling of multi-state sign glosses for
  continuous sign language recognition.
\newblock In {\em Computer Vision--ECCV 2020: 16th European Conference,
  Glasgow, UK, August 23--28, 2020, Proceedings, Part XVI 16}, pages 172--186.
  Springer, 2020.

\bibitem[\protect\citeauthoryear{Pu \bgroup \em et al.\egroup
  }{2019}]{pu2019iterative}
Junfu Pu, Wengang Zhou, and Houqiang Li.
\newblock Iterative alignment network for continuous sign language recognition.
\newblock In {\em Proceedings of the IEEE/CVF conference on computer vision and
  pattern recognition}, pages 4165--4174, 2019.

\bibitem[\protect\citeauthoryear{Radford \bgroup \em et al.\egroup
  }{2021}]{radford2021learning}
Alec Radford, Jong~Wook Kim, Chris Hallacy, Aditya Ramesh, Gabriel Goh,
  Sandhini Agarwal, Girish Sastry, Amanda Askell, Pamela Mishkin, Jack Clark,
  et~al.
\newblock Learning transferable visual models from natural language
  supervision.
\newblock In {\em International conference on machine learning}, pages
  8748--8763. PMLR, 2021.

\bibitem[\protect\citeauthoryear{Sarto \bgroup \em et al.\egroup
  }{2023}]{sarto2023positive}
Sara Sarto, Manuele Barraco, Marcella Cornia, Lorenzo Baraldi, and Rita
  Cucchiara.
\newblock Positive-augmented contrastive learning for image and video
  captioning evaluation.
\newblock In {\em Proceedings of the IEEE/CVF Conference on Computer Vision and
  Pattern Recognition}, pages 6914--6924, 2023.

\bibitem[\protect\citeauthoryear{Teed and Deng}{2020}]{teed2020raft}
Zachary Teed and Jia Deng.
\newblock Raft: Recurrent all-pairs field transforms for optical flow.
\newblock In {\em Computer Vision--ECCV 2020: 16th European Conference,
  Glasgow, UK, August 23--28, 2020, Proceedings, Part II 16}, pages 402--419.
  Springer, 2020.

\bibitem[\protect\citeauthoryear{Ur~Rehman \bgroup \em et al.\egroup
  }{2023}]{ur2023use}
Atiq Ur~Rehman, Samir~Brahim Belhaouari, Md~Alamgir Kabir, and Adnan Khan.
\newblock On the use of deep learning for video classification.
\newblock {\em Applied Sciences}, 13(3):2007, 2023.

\bibitem[\protect\citeauthoryear{Wang \bgroup \em et al.\egroup
  }{2020}]{wang2020deep}
Jingdong Wang, Ke~Sun, Tianheng Cheng, Borui Jiang, Chaorui Deng, Yang Zhao,
  Dong Liu, Yadong Mu, Mingkui Tan, Xinggang Wang, et~al.
\newblock Deep high-resolution representation learning for visual recognition.
\newblock {\em IEEE transactions on pattern analysis and machine intelligence},
  43(10):3349--3364, 2020.

\bibitem[\protect\citeauthoryear{Wei and Chen}{2023}]{wei2023improving}
Fangyun Wei and Yutong Chen.
\newblock Improving continuous sign language recognition with cross-lingual
  signs.
\newblock {\em arXiv preprint arXiv:2308.10809}, 2023.

\bibitem[\protect\citeauthoryear{Xie \bgroup \em et al.\egroup
  }{2018}]{xie2018rethinking}
Saining Xie, Chen Sun, Jonathan Huang, Zhuowen Tu, and Kevin Murphy.
\newblock Rethinking spatiotemporal feature learning: Speed-accuracy trade-offs
  in video classification.
\newblock In {\em Proceedings of the European conference on computer vision
  (ECCV)}, pages 305--321, 2018.

\bibitem[\protect\citeauthoryear{Xing \bgroup \em et al.\egroup
  }{2023}]{xing2023svformer}
Zhen Xing, Qi~Dai, Han Hu, Jingjing Chen, Zuxuan Wu, and Yu-Gang Jiang.
\newblock Svformer: Semi-supervised video transformer for action recognition.
\newblock In {\em Proceedings of the IEEE/CVF Conference on Computer Vision and
  Pattern Recognition}, pages 18816--18826, 2023.

\bibitem[\protect\citeauthoryear{Yang \bgroup \em et al.\egroup
  }{2020}]{yang2020temporal}
Ceyuan Yang, Yinghao Xu, Jianping Shi, Bo~Dai, and Bolei Zhou.
\newblock Temporal pyramid network for action recognition.
\newblock In {\em Proceedings of the IEEE/CVF conference on computer vision and
  pattern recognition}, pages 591--600, 2020.

\bibitem[\protect\citeauthoryear{Yang \bgroup \em et al.\egroup
  }{2023}]{yang2023vid2seq}
Antoine Yang, Arsha Nagrani, Paul~Hongsuck Seo, Antoine Miech, Jordi
  Pont-Tuset, Ivan Laptev, Josef Sivic, and Cordelia Schmid.
\newblock Vid2seq: Large-scale pretraining of a visual language model for dense
  video captioning.
\newblock In {\em Proceedings of the IEEE/CVF Conference on Computer Vision and
  Pattern Recognition}, pages 10714--10726, 2023.

\bibitem[\protect\citeauthoryear{Zhang \bgroup \em et al.\egroup
  }{2022}]{zhang2022tip}
Renrui Zhang, Wei Zhang, Rongyao Fang, Peng Gao, Kunchang Li, Jifeng Dai,
  Yu~Qiao, and Hongsheng Li.
\newblock Tip-adapter: Training-free adaption of clip for few-shot
  classification.
\newblock In {\em European Conference on Computer Vision}, pages 493--510.
  Springer, 2022.

\bibitem[\protect\citeauthoryear{Zheng \bgroup \em et al.\egroup
  }{2023}]{zheng2023cvt}
Jiangbin Zheng, Yile Wang, Cheng Tan, Siyuan Li, Ge~Wang, Jun Xia, Yidong Chen,
  and Stan~Z Li.
\newblock Cvt-slr: Contrastive visual-textual transformation for sign language
  recognition with variational alignment.
\newblock In {\em Proceedings of the IEEE/CVF Conference on Computer Vision and
  Pattern Recognition}, pages 23141--23150, 2023.

\bibitem[\protect\citeauthoryear{Zhou \bgroup \em et al.\egroup
  }{2021a}]{zhou2021improving}
Hao Zhou, Wengang Zhou, Weizhen Qi, Junfu Pu, and Houqiang Li.
\newblock Improving sign language translation with monolingual data by sign
  back-translation.
\newblock In {\em Proceedings of the IEEE/CVF Conference on Computer Vision and
  Pattern Recognition}, pages 1316--1325, 2021.

\bibitem[\protect\citeauthoryear{Zhou \bgroup \em et al.\egroup
  }{2021b}]{zhou2021spatial}
Hao Zhou, Wengang Zhou, Yun Zhou, and Houqiang Li.
\newblock Spatial-temporal multi-cue network for sign language recognition and
  translation.
\newblock {\em IEEE Transactions on Multimedia}, 24:768--779, 2021.

\bibitem[\protect\citeauthoryear{Zhou \bgroup \em et al.\egroup
  }{2023}]{zhou2023gloss}
Benjia Zhou, Zhigang Chen, Albert Clap{\'e}s, Jun Wan, Yanyan Liang, Sergio
  Escalera, Zhen Lei, and Du~Zhang.
\newblock Gloss-free sign language translation: Improving from visual-language
  pretraining.
\newblock {\em arXiv preprint arXiv:2307.14768}, 2023.

\bibitem[\protect\citeauthoryear{Zuo and Mak}{2022}]{zuo2022c2slr}
Ronglai Zuo and Brian Mak.
\newblock C2slr: Consistency-enhanced continuous sign language recognition.
\newblock In {\em Proceedings of the IEEE/CVF Conference on Computer Vision and
  Pattern Recognition}, pages 5131--5140, 2022.

\end{thebibliography}
\appendix

\section{Study of Different Fusion Methods.}~\label{section:appendix2}
We investigated the effectiveness of our fusion methods in this section. 
Results in Table \ref{table: sup2} clearly demonstrate that each fusion method contributed to a reduction in error rates for SLR. And it was observed that the MLP-based fusion method demonstrated the highest effectiveness among the fusion methods employed. We should note that in this ablation study, we exclude the visual-textual alignment approach.

MLP-based fusion method is a simple yet effective method, which contains two linear layers, each layer followed by a GELU activation function except for the last one. The operation of the fusion module can be represented as follows: 
\begin{align}
    f^{m} = \text{MLP} (f^{v} \copyright f^{k} \copyright f^{o}),\quad f^{v'}= f^m + f^v,
\end{align} 
\begin{align}
    f^{k'} = f^m + f^k, \quad f^{o'} = f^m + f^o,
\end{align} 
where \text{MLP} represent the MLP function, $f^m$ is the feature output from our fusion module, $f^v$, $f^k$, $f^o$ are features input to the fusion module, and $f^{v'}$, $f^{k'}$, $f^{o'}$ are enhanced features input to the next stage. The same meanings of the symbols are below.

For the convolution-based fusion method, we describe the process as:
\begin{align}
    f^m = \text{Conv}(f^{v} \copyright f^{k} \copyright f^{o}), \quad f^{v'}= f^m + f^v,
\end{align}
\begin{align}
    f^{k'} = f^m + f^k, \quad f^{o'} = f^m + f^o,
\end{align} 
where \text{Conv} is a $3 \times 3 \times 3$ convolutional layer.

In terms of the attention-based fusion method, the operation of the fusion module can be represented as: 
\begin{align}
    f^{v'} = f^{v} + \text{Attn}(f^{v}, f^{k}) + \text{Attn}(f^{v}, f^{o}),
\end{align} 
\begin{align}
    f^{k'} = f^{k} + \text{Attn}(f^{k}, f^{v}) + \text{Attn}(f^{k}, f^{o}), 
\end{align} 
\begin{align}
    f^{o'} = f^{o} + \text{Attn}(f^{o}, f^{v}) + \text{Attn}(f^{o}, f^{k}),
\end{align} 
where \text{Attn} represents the cross-attention function.

\section{More Experiments on the Phoenix-2014T and CSL-Daily Datasets.}~\label{section:appendix1}
We performed individual testing of each encoder of our SignVTCL on the Phoenix-2014T and CSL-Daily datasets, as shown in Table \ref{table: sup1}. Simultaneously, the effectiveness of the fusion module was also validated.

\section{More Visualizations}~\label{section:appendix3}
\begin{figure*}[!]
    \centering
        \includegraphics[width=\linewidth]{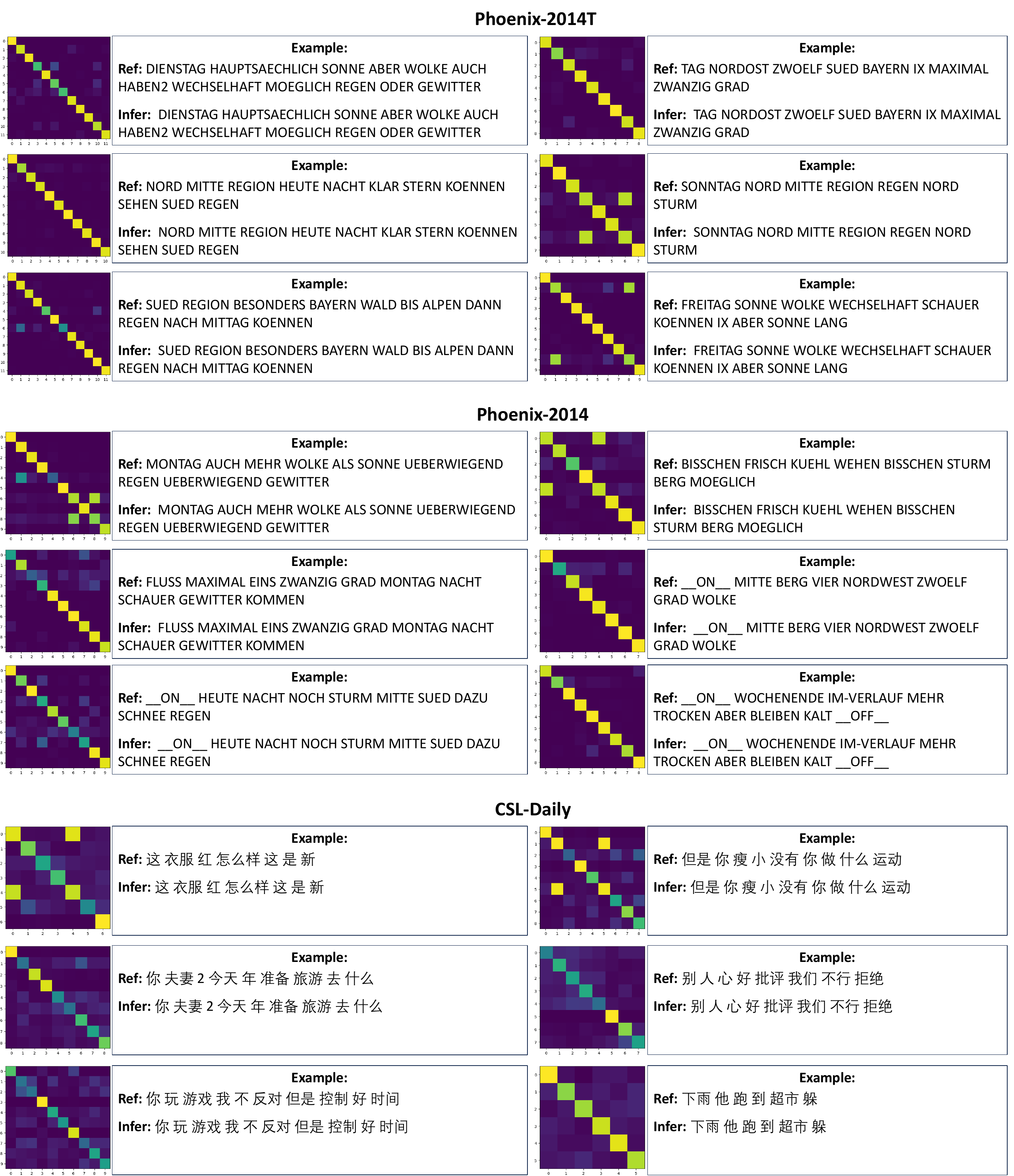}
    \caption{\textbf{Visualization of Gloss-Level Alignment Similarity Matrix and Predicted Glosses on the Phoenix-2014T, Phoenix-2014, and CSL-Daily Datasets.}}
    \vspace{-10pt}
    \label{fig:vis1}
\end{figure*}

\begin{figure*}[!]
    \centering
        \includegraphics[width=\linewidth]{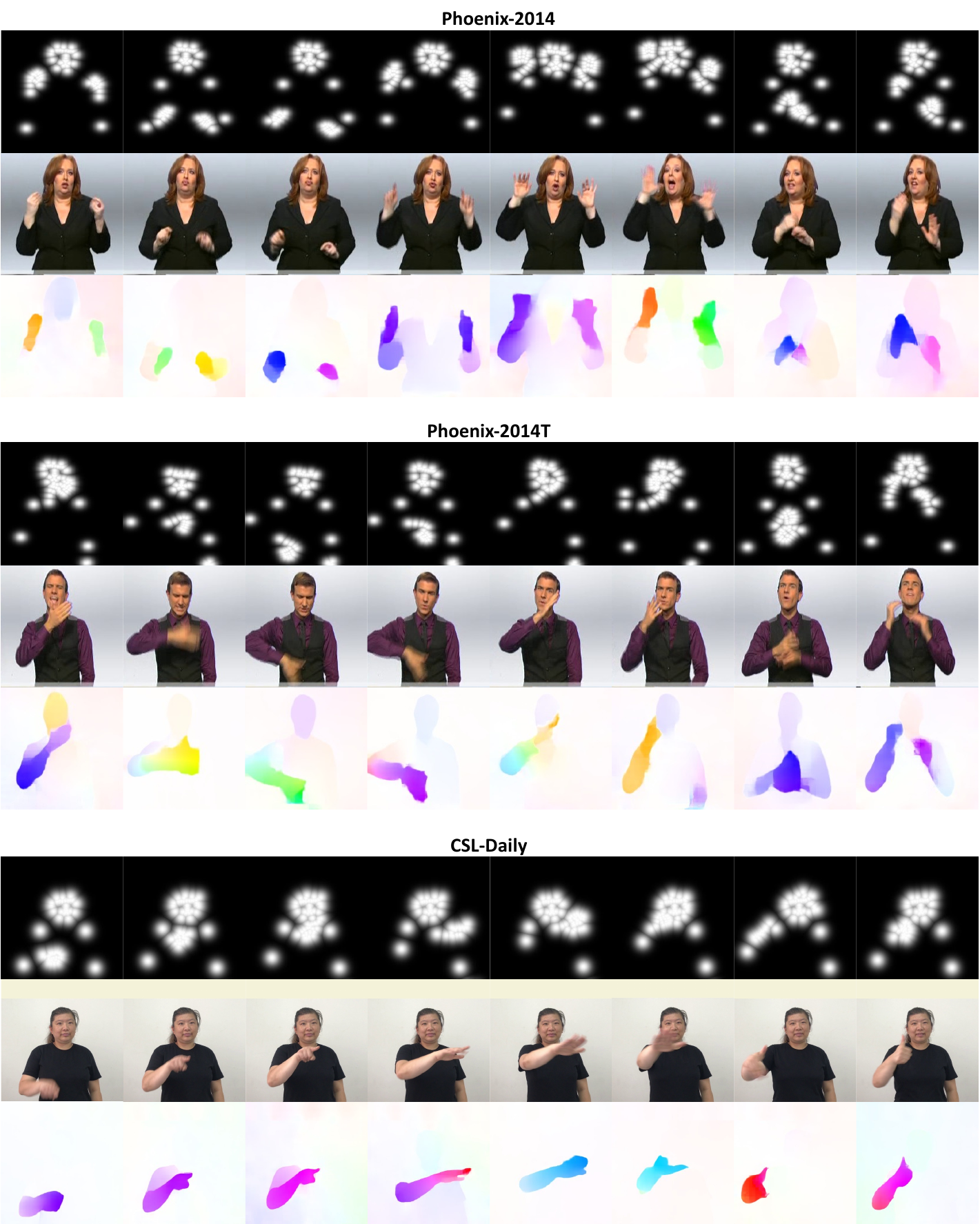}
    \caption{\textbf{Visualization of Three Input Modalities Data on the Phoenix-2014T, Phoenix-2014, and CSL-Daily Datasets.} For each example, the first line is keypoints, the second line is video, and the third line is optical flow.}
    \vspace{-10pt}
    \label{fig:vis2}
\end{figure*}

To verify that our gloss-level alignment plays a pivotal role in efficiently aligning visual and textual features, we show more similarity matrices and predicted glosses in Figure \ref{fig:vis1}.
Additionally, we also visualized the input data in three modalities, as shown in Figure \ref{fig:vis2}. Keypoints can eliminate the variations in signing styles among individuals and optical flow can effectively capture dynamic body parts movements.
Specifically, given as a set of keypoints, they are processed by a Gaussian function to generate heatmaps $x^k \in \mathcal{R}^{T\times H\times W\times K}$. $x^k_{(t,i,j,k)} = \text{exp}(-[(i-x^k_t)^2+(j-y^k_t)^2]/2\sigma^2)$, where $(x^k_t,y^k_t)$ denotes the coordinates of the $k$-th keypoint in the $t$-th frame, and $\sigma$ is a scale controller.

\begin{table}[!]
    \centering
    \setlength{\tabcolsep}{2mm}{
    \begin{tabular}{{ccc|cc}}
        \toprule[1.5pt]
        \multirow{2}*{Conv} & \multirow{2}*{Attn} & \multirow{2}*{MLP} & \multicolumn{2}{c}{WER} \\
        \cmidrule(lr){4-5}&&& Dev (\%) &Test (\%)\\
        \midrule[0.5pt]
         \ding{51}&  &   & 18.4& 18.8\\
         &  \ding{51} &   & 18.6& 19.0\\
         \rowcolor{gray!20} &   &  \ding{51} & \textbf{18.3}& \textbf{18.6}\\
        \bottomrule[1.5pt]
    \end{tabular}
    }
    \caption{\textbf{Ablation Study for the Effectiveness of Each Fusion Method on the Phoenix-2014 dataset.} `Conv', `Attn', `MLP' are convolution-based, attention-based and MLP-based method.}
    \vspace{-1pt}
    \label{table: sup2}
\end{table}

\begin{table}[!]
    \centering
    \setlength{\tabcolsep}{1.5mm}{
    \begin{tabular}{{l|cccc|cc}}
        \toprule[1.5pt]
        \multirow{2}*{Dataset}& \multirow{2}*{VE}& \multirow{2}*{KE} & \multirow{2}*{OE} & \multirow{2}*{FM} & \multicolumn{2}{c}{WER} \\
        \cmidrule(lr){6-7} & & & & & Dev (\%) &Test (\%)\\
        \midrule[0.5pt]
         \multirow4*{Phoenix-2014T}&\ding{51}&  &  & & 20.6& 21.4\\
         & &  \ding{51} &  & & 26.3& 26.2\\
         & &   &  \ding{51} & & 38.1& 37.7\\
         & \ding{51} &  \ding{51} & \ding{51} & & 19.5& 20.2\\
         & \cellcolor{gray!20}\ding{51} &   \cellcolor{gray!20}\ding{51} &  \cellcolor{gray!20}\ding{51} &  \cellcolor{gray!20}\ding{51} &  \cellcolor{gray!20}\textbf{18.1}&  \cellcolor{gray!20}\textbf{19.0}\\
        \midrule[0.5pt]
         \multirow4*{CSL-Daily}&\ding{51}&  &  & & 28.1& 27.9\\
         & &  \ding{51} &  & & 34.7& 33.8\\
         & &   &  \ding{51} & & 36.7& 36.1\\
         & \ding{51} &  \ding{51} & \ding{51} & & 26.8& 26.6\\
         & \cellcolor{gray!20}\ding{51} &   \cellcolor{gray!20}\ding{51} &  \cellcolor{gray!20}\ding{51} &  \cellcolor{gray!20}\ding{51} &  \cellcolor{gray!20}\textbf{25.4}&  \cellcolor{gray!20}\textbf{25.0}\\
        \bottomrule[1.5pt]
    \end{tabular}
    }
    \caption{\textbf{Ablation Study for the Effectiveness of Each Encoder and Fusion Module on the Phoenix-2014T and CSL-Daily Datasets.} `VE', `KE', `OE' are video encoder, keypoint encoder and optical flow encoder, respectively. `FM' denotes the fusion module.} 
    \vspace{-5pt}
    \label{table: sup1}
\end{table}


\end{document}